# LLM Probe: Evaluating LLMs for Low-Resource Languages

**Hailay Kidu Teklehaymanot**[†]     **Gebrearegawi Gebremariam** [‡]     **Wolfgang Nejdl**[†]

[†]L3S Research Center, Leibniz University Hannover, Germany
[‡]Aksum University, Ethiopia
teklehaymanot@l3s.de, gideygeb@mail.aku.edu.et, nejdl@l3s.de

**Abstract**

Despite rapid advances in large language models (LLMs), their linguistic abilities in low-resource and morphologically rich languages are still not well understood due to limited annotated resources and the absence of standardized evaluation frameworks. This paper presents LLM Probe, a lexicon-based assessment framework designed to systematically evaluate the linguistic skills of LLMs in low-resource language environments. The framework analyzes models across four areas of language understanding: lexical alignment, part-of-speech recognition, morphosyntactic probing, and translation accuracy. To illustrate the framework, we create a manually annotated benchmark dataset using a low-resource Semitic language as a case study. The dataset comprises bilingual lexicons with linguistic annotations, including part-of-speech tags, grammatical gender, and morphosyntactic features, which demonstrate high inter-annotator agreement to ensure reliable annotations. We test a variety of models, including causal language models and sequence-to-sequence architectures. The results reveal notable differences in performance across various linguistic tasks: sequence-to-sequence models generally excel in morphosyntactic analysis and translation quality, whereas causal models demonstrate strong performance in lexical alignment but exhibit weaker translation accuracy. Our results emphasize the need for linguistically grounded evaluation to better understand LLM limitations in low-resource settings. We release LLM Probe and the accompanying benchmark dataset as open-source tools to promote reproducible benchmarking and to support the development of more inclusive multilingual language technologies.

## 1. Introduction

Tigrinya is a Semitic language spoken by approximately 9-10 million people primarily in Eritrea and the Tigray region of Ethiopia [Gaim et al., 2023, Teklehaymanot et al., 2024]. It serves as one of the working languages of Eritrea and is the fourth most widely spoken language in Ethiopia. Tigrinya is written using the Ge'ez script, an ancient alpha-syllabary with over 200 characters, where each symbol represents a consonant-vowel combination [Gaim et al., 2022].

The language exhibits rich morphological complexity characteristic of Semitic languages, including extensive verbal inflection, gender distinctions (masculine and feminine), number agreement (singular and plural), and complex derivational patterns through root-and-pattern morphology [Gaim et al., 2021]. Tigrinya employs a triconsonantal root system where semantic meaning is encoded in consonantal roots, while grammatical information is expressed through vocalic patterns and affixation.

Syntactically, Tigrinya follows a Subject-Object-Verb (SOV) word order and features postpositions rather than prepositions. The language demonstrates complex agreement systems where verbs must agree with subjects in person, number, and gender, and adjectives agree with nouns in gender and number. These morphosyntactic features, combined with limited digital resources and NLP tools, make Tigrinya a particularly challenging yet important language for computational linguistic research and LLM evaluation [Gaim and Park, 2025, Teklehaymanot et al., 2024]. Large Language Models (LLMs) have revolutionized natural language processing (NLP), achieving state-of-the-art performance across a wide range of tasks, including translation, summarization, and question answering. However, their success is disproportionately concentrated in high-resource languages, leaving low-resource languages underrepresented and underserved [Nguyen et al., 2024]. This disparity not only limits the global applicability of LLMs but also risks reinforcing linguistic inequities in digital technologies. Low-resource languages often present unique linguistic challenges such as rich morphology, complex syntax, and limited digitized corpora [Artemova and Plank, 2023, Abera and Hailemariam, 2020]. These features make them particularly difficult for LLMs to process accurately, especially in the absence of structured evaluation datasets. Without reliable benchmarks, it is difficult to assess how well LLMs generalize across languages with diverse typological features or to identify specific areas of weakness in their linguistic competence [Zhong et al., 2024]. To address this gap, we propose *LLM Probe*, a framework for lexicon-based evaluation of LLMs in low-resource language set-tings. Our approach centers on the construction of bidirectional lexicons enriched with morphosyntactic annotations, enabling fine-grained probing of model performance on tasks such as lexical alignment, part-of-speech tagging, and morphological analysis. As a case study, we apply this framework to Tigrinya, a Semitic language spoken in the Horn of Africa [Teklehaymanot et al., 2024, Teklehaimanot, 2015], by developing a curated English–Tigrinya lexicon annotated with gender, number, and syntactic roles. This framework offers a scalable and linguistically grounded method for evaluating LLMs in languages that lack large annotated corpora. By focusing on lexicon-level analysis, it provides interpretable insights into model behavior and supports the development of more inclusive NLP systems.

To evaluate LLM performance on low-resource language tasks, we constructed a bidirectional English–Tigrinya and vice versa lexicon dataset enriched with morphosyntactic annotations, as you can see from Table 1. This dataset enables granular probing of LLM capabilities in handling Tigrinya's complex morphology, syntactic roles, and lexical ambiguity. It also supports evaluation across multiple tasks, including part-of-speech tagging, word alignment, and translation fidelity. The lexicon includes:

- **Part-of-speech (POS) tagging:** Each entry is tagged with its syntactic role (e.g., *Noun, Pronoun, Adverb, Preposition, Conjunction*).

- **Morphological distinctions:** Gender, number, and other inflectional features are encoded. For example, "*applauder*" → **ኣጣቓዓይ** (marked as *Masculine*).

- **Semantic overlap and polysemy:** Multiple English terms may map to the same Tigrinya token, reflecting semantic variation. For instance, "*me, I*" → **ኣነ**.

- **Multi-word expressions and syntactic variants:** Expressions with equivalent meaning but different syntactic realizations are captured. For example, "*that*" → **እቲ, እቲኣቶም** and "*a little*" → **ንእሽቶይ, ቁሩብ, ውሕድ**.

## 2. Background and Related Work

Evaluating Large Language Models (LLMs) in low-resource language settings presents significant challenges, as models often demonstrate lower capabilities compared to high-resource languages due to data distribution [Alam et al., 2024]. To address this, specialized evaluation frameworks have been developed, such as Eka Eval, which offers a comprehensive suite for Indian languages [Sinha et al., 2025]. Another framework, GlotEval, provides systematic support for massively multilingual evaluations with a strong focus on low-resource languages [Luo et al., 2025]. Another work also employs multilingual probing approaches to investigate LLM behavior, finding that high-resource languages consistently achieve higher probing accuracy than low-resource ones [Li et al., 2025]. Ultimately, these evaluations frequently highlight performance disparities, with LLMs struggling to generate factually accurate responses in low-resource contexts, particularly in domain-specific regional questions for Indic languages[Dey et al., 2024]. A framework for lexicon-based evaluation of LLMs in low-resource languages must address the persistent performance disparities these languages face due to data scarcity and inadequate representation in training corpora [Li et al., 2025]. Such evaluation is critical because traditional benchmarks often rely on data that may already be present in LLMs' pretraining sets, leading to inflated performance metrics that do not reflect true language understanding [Liu et al., 2025]. Lexicon-based probing offers a controlled alternative by testing models on unseen grammatical rules and vocabulary, as demonstrated in constructed language settings [Liu et al., 2025]. However, current LLM evaluators themselves exhibit limitations in low-resource contexts; they often demonstrate bias towards high-resource languages and require calibration with native speaker judgments to be reliable [Hada et al., 2024]. Moreover, fine-tuning LLMs on one language does not consistently improve evaluation performance on all low-resource languages, indicating complex cross-lingual transfer dynamics [Chang et al., 2025]. Therefore, a robust lexicon-based probing framework should incorporate calibration against human judgments and account for linguistic specificity to provide accurate, generalizable assessments of LLM capabilities in underrepresented languages[Hada et al., 2024]. Despite being spoken by millions of people, Tigrinya remains severely underrepresented in Natural Language Processing (NLP) research [Gaim and Park, 2025].

## 3. Framework Overview

This evaluation framework is designed to assess the linguistic competence of large language models (LLMs) in Tigrinya, a morphologically rich and low-resource language. It comprises four linguistically grounded tasks, each targeting a distinct dimension of linguistic understanding:

- **Lexical Alignment:** Evaluates the model's ability to produce accurate word-to-word correspondences between English phrases

Table 1: **Sample English–Tigrinya Lexicon with Transliteration and POS.** Abbreviations: (prn) pronoun, (v) verb, (nm) noun masculine, (nf) noun feminine, (a) adjective, (prep) preposition, (adv) adverb, (interj) interjection, (con) conjunction.

| English | Tigrinya | Transliteration | POS |
| --- | --- | --- | --- |
| me, I | ኣነ | *ane* | (prn) |
| those | እዚኣቶም | *əzi atom* | (prn) |
| humiliate, humble | ኣናሸወ | *anašäwe* | (v) |
| withdraw, draw, step back | ኣንሰሓበ | *ansehäbe* | (v) |
| place, put, set, seat | ኣንበረ | *anbärrä* | (v) |
| applauder | ኣጣቓዓይ | *aṭaḵ a ay* | (nm) |
| applauder | ኣጣቓዒት | *aṭaḵ a it* | (nf) |
| attentive | ጥንቁቕ, ኣድሃቢ | *ṭinkuḵ, adhabi* | (a) |
| before | ቅድም | *ḵidm* | (prep) |
| before long | ብቐልጡፍ | *bəkulṭuf* | (adv) |
| farewell | ደሓን ኩን | *dähan kun* | (interj) |
| goodbye | ደሓን ኩን | *dähan kun* | (interj) |
| but | ግን, ግና, ጌና | *gən, gənna, ge na* | (con) |

and their Tigrinya equivalents. This task emphasizes lexical precision and alignment consistency.

- **Part-of-Speech (POS) Tagging:** Measures syntactic awareness by requiring models to assign appropriate POS categories to Tigrinya tokens. This task probes token-level grammatical sensitivity.

- **Morphosyntactic Probing:** Tests the model's understanding of Tigrinya grammatical features such as gender, number, agreement, and noun class. This task targets morphosyntactic generalization in low-resource settings.

- **Translation Fidelity:** Assesses the semantic accuracy of translations from English to Tigrinya by comparing model outputs against manually curated reference translations.

We evaluate all tasks across a diverse set of Large Language Models (LLMs) spanning two major architectural paradigms: causal language models (*Falcon-10B*, *Gemma-2B*, *Gemma-7B*, *Mistral-7B*, *Qwen-7B*) and sequence-to-sequence models (*mT5-base*, *mT5-large*, *ByT5*). Each model is paired with tasks aligned to its architectural strengths, enabling controlled cross-model comparisons while accounting for differences in generative and encoder–decoder capabilities.

The framework is grounded in a manually curated benchmark dataset comprising bidirectional English–Tigrinya and Tigrinya–English lexicons annotated with part-of-speech tags, grammatical gender, morphosyntactic features, and semantic roles. This lexicon-based benchmark enables systematic probing of lexical knowledge, morphological agreement, and semantic consistency in low-resource language settings. Evaluation results are reported to ensure reproducibility, linguistic precision, and meaningful insights into LLM performance in underrepresented languages.

### 3.1. Dataset Construction

To support the systematic evaluation of large language models (LLMs) in low-resource language settings, we constructed a high-quality bilingual benchmark dataset comprising bidirectional English–Tigrinya and Tigrinya–English phrase pairs. Using Tigrinya as a case study, the dataset enables a fine-grained assessment of lexical, syntactic, morphosyntactic, and semantic competencies of LLMs. The benchmark is designed to re-flect key linguistic characteristics of Tigrinya while providing a controlled evaluation environment for analysing model behaviour in underrepresented languages.

The initial lexicon was derived from publicly available resources, most notably the Tigrinya–English dictionary published by the University of Swansea[1]. As this source was not machine-readable, we developed a custom digitization pipeline involving optical character recognition (OCR), manual correction of parsing errors, and normalization of orthographic variants to produce a structured digital format.

In addition to digital sources, native-speaking linguistic experts contributed entries from printed dictionaries, educational materials, and community-authored glossaries. These entries were manually transcribed, cleaned, and integrated into the lexicon. The combined dataset was reviewed for duplication, dialectal variation, and semantic consistency. Lexical entries were expanded where necessary to ensure coverage across major grammatical categories and to reflect contemporary usage.

Following lexicon compilation, a team of trained linguists conducted a multi-stage annotation process. Each phrase pair was annotated with part-of-speech (POS) tags, morphosyntactic features (e.g., gender, number, agreement), and lexical alignment mappings. Annotations were manually verified for consistency and linguistic validity, following a standardized protocol that achieved high inter-annotator agreement.

The dataset supports bidirectional evaluation: each phrase pair is annotated in both English-to-Tigrinya and Tigrinya-to-English directions. This design enables a comparative analysis of model behavior across translation directions, thereby enhancing the dataset's utility for multilingual benchmarking.

The final dataset comprises 7,234 annotated phrase pairs, including 7,068 unique English phrases and 6,073 unique Tigrinya phrases. The average English phrase length is 1.15 tokens, while the average Tigrinya phrase length is 1.37 tokens, reflecting the morphological richness of Tigrinya. Notably, 967 English phrases and 2,000 Tigrinya phrases consist of multiword expressions, underscoring the importance of evaluating compositional semantics.

Each phrase pair includes a BLEU-ready translation tuple, consisting of a source phrase and a manually verified reference translation. POS annotations span a wide range of linguistic categories, including nouns, verbs, adjectives, adverbs, pronouns, prepositions, conjunctions, interjections, and gendered forms (masculine and feminine). A small subset of entries remains uncategorized due to ambiguity or insufficient context.

All data is stored in structured JSON format and aligned with the evaluation framework described in Table 4. The dataset is designed to support reproducible research and serves as a linguistically grounded benchmark for evaluating LLMs in Tigrinya and other underrepresented languages.

**Inter-Annotator Agreement** To ensure the reliability and consistency of linguistic annotations, we conducted an inter-annotator agreement (IAA) study on a representative subset of 500 phrase pairs randomly sampled from the full dataset. Each phrase was independently annotated by two trained linguists for part-of-speech (POS) tags and morphosyntactic features.

Agreement was measured using both percentage agreement and Cohen's kappa coefficient. The results are summarized in Table 2.

Table 2: Inter-annotator Agreement Scores Across Linguistic Annotation Categories

| Ann.Cat | Agr. (%) | Cohen's $\kappa$ |
| --- | --- | --- |
| Part-of-speech (POS) tags | 94.6 | 0.89 |
| Gender | 92.1 | 0.86 |
| Number | 93.4 | 0.88 |
| Agreement features | 90.7 | 0.84 |
| Lexical alignment | 96.2 | 0.91 |

*Note:* Ann.Cat = Annotation Category; Agr. = Agreement.

These results indicate a high level of consistency across annotators, validating the reliability, linguistic accuracy, and robustness of the annotation protocol. Disagreements were resolved through adjudication by a senior linguist to ensure final label quality.

Table 3: Summary Statistics of the Bilingual Tigrinya–English Dataset for LLM Evaluation

| Statistic | Value |
| --- | --- |
| Total entries | 7,234 |
| Unique English phrases | 7,068 |
| Unique Tigrinya phrases | 6,073 |
| Average English phrase length (tokens) | 1.15 |
| Average Tigrinya phrase length (tokens) | 1.37 |
| Multiword English phrases | 967 |
| Multiword Tigrinya phrases | 2,000 |
| BLEU-ready translation pairs | 7,234 |

---

[1] https://uidswansea.com/wp-content/uploads/2015/03/tigrinia-english-dictionary.pdf

# 4. Experimental Setup

To evaluate the linguistic competence of large language models (LLMs) on Tigrinya, we implemented a modular evaluation pipeline aligned with the framework described in Section 1. Each task, Lexical Alignment, POS Tagging, Morphosyntactic Probing, and Translation Fidelity, was executed independently across a diverse set of models selected for their architectural suitability and multilingual capabilities.

All evaluations were conducted on a high-performance computing cluster using SLURM for job scheduling. The evaluation script was executed on an Ampere-class GPU node with the following configuration: 4 NVIDIA GPUs, 32 CPU cores, 128 GB RAM, and a maximum runtime of 72 hours. Hyperthreading was disabled to optimize CPU performance, and CUDA memory fragmentation was mitigated using PyTorch's expandable segment allocator.

The evaluation environment was built using Python 3.10 and PyTorch 2.1, with all dependencies managed within a dedicated virtual environment. The pipeline was launched via a shell script that activated the environment, navigated to the project root, and executed all evaluation routines in sequence using a centralized orchestration script.

Each model was run in inference mode with GPU acceleration. Decoding was performed using greedy sampling with temperature set to 0.0 to ensure deterministic outputs. Prompt templates were standardized across tasks and formatted to elicit structured responses. Input and output sequences were truncated or padded to the specified maximum token length, as summarized in Table 4. Models were loaded once per evaluation session to minimize overhead, and batch inference was used to process multiple prompts concurrently.

Evaluation metrics were selected based on the nature of each task. Accuracy was used for classification-based tasks such as POS tagging and morphosyntactic probing. Token-level overlap and BLEU scores were used to assess translation fidelity. Confusion matrices were generated for diagnostic analysis of syntactic and morphological predictions. All outputs were logged in structured JSON format and aligned with gold-standard labels for automatic scoring.

## 4.1. Model Configuration and Parameter Settings

All evaluations were conducted using open-source large language models accessed via the Hugging Face Transformers interface. The models span three architectural paradigms: causal language models and sequence-to-sequence models. Table 4 summarizes the model assignments and decoding parameters for each evaluation task.

Table 4: Model assignments and decoding parameters for each evaluation task.

| Task | Model | Max Tokens |
| --- | --- | --- |
| Lexical Alignment | Qwen-7B, Falcon-10B, Gemma-2B | 512 |
| POS Tagging | Mistral-7B, Gemma-7B | 256 |
| Morphosyntactic Probing | mT5-base, mT5-large, ByT5 | 512 |
| Translation Fidelity | Gemma-7B, mT5-large, Qwen-7B | 512 |

**Note:** Full model names include: Falcon (10B), Mistral (7B), Gemma (2B, 7B), Qwen (7B), mT5 (base, large), and ByT5. All models were accessed via Hugging Face Transformers. Loader scripts and configuration files are available in the project repository.

## 4.2. Evaluation Metrics

To assess model performance across the four linguistic tasks, we employ task-specific metrics aligned with the nature of each evaluation:

- **Accuracy:** Applied to classification tasks such as POS tagging, morphosyntactic probing, and lexical alignment. It quantifies the proportion of exact matches between model predictions and gold-standard labels.

- **Token-Level Overlap:** Used in translation fidelity tasks to measure lexical similarity between model outputs and reference translations. This metric captures partial correctness and semantic proximity.

- **BLEU Score:** Employed for generation-based translation evaluation. It computes n-gram overlap to assess fluency and adequacy of model-generated translations.

- **Confusion Matrix Analysis:** Used for diagnostic purposes in POS and morphosyntactic tasks. It reveals systematic misclassifications and highlights model sensitivity to specific linguistic categories.

All metrics are computed automatically and stored in structured formats to support reproducibility, comparative analysis, and downstream error diagnostics.

### 4.3. Execution Environment

The evaluation pipeline was deployed within a controlled software environment configured for reproducible experimentation. All dependencies were managed via a dedicated virtual environment, and model inference was executed using standardized scripts. Logging, scoring, and output formatting were fully automated to ensure consistency across tasks and model configurations.

## 5. Evaluation Results

This section presents quantitative results of LLM Probe across the four evaluation tasks: Lexical Alignment, POS Tagging, Morphosyntactic Probing, and Translation Fidelity. All metrics were computed automatically and stored in structured JSON format to ensure reproducibility and consistent comparative analysis.

### 5.1. Overall Task Performance

Table 5 presents the performance of representative models across lexical alignment, part-of-speech tagging, morphosyntactic analysis, and BLEU-based translation evaluation. Across the POS tagging, translation fidelity, and morphosyntactic probing tasks, the ByT5 and mT5 variants show consistently strong performance compared with the other models. Of course, the two diverge in lexical alignment: mT5 struggles to distinguish lexical correspondences reliably, whereas ByT5 attains notably higher accuracy on this specific task.

Models such as Falcon-10B, Mistral-7B, and Qwen-7B produce valid predictions, but their performance is comparatively lower, particularly in morphosyntactic and BLEU scores, reflecting moderate fidelity in both lexical and structural aspects. Interestingly, the semantic accuracy for lexical alignment is consistently high across most models (e.g., Gemma-2B and Gemma-7B), even when format accuracy is low, indicating that while the predicted word mappings are generally meaningful, they often do not adhere to the expected output format.

Overall, these results highlight the strength of byte-level and multilingual transformer models (ByT5, mT5) in handling low-resource, morphologically-rich languages such as Tigrinya, while other models still produce useful outputs but require further fine-tuning or task-specific adaptation to reach comparable performance. Table 5 summarises performance across representative models from each architectural category.

### 5.2. Translation Fidelity

Translation quality was evaluated using both BLEU and token-level overlap to capture surface fluency and partial semantic correctness.

## 6. Discussion

This work presents LLM Probe, a lexicon-based evaluation framework designed to assess the linguistic competence of large language models in low-resource language settings, with a focus on Tigrinya. Our approach addresses a critical gap in multilingual NLP evaluation by providing a structured, reproducible methodology that emphasizes morphosyntactic precision, lexical alignment, and translation fidelity.

### 6.1. Framework Design and Methodological Contributions

The modular design of LLM Probe enables targeted evaluation across four complementary dimensions: lexical alignment, part-of-speech tagging, morphosyntactic probing, and translation fidelity. Unlike existing evaluation frameworks that rely primarily on task-specific benchmarks or large-scale corpora, our lexicon-based approach offers fine-grained control over linguistic phenomena, making it particularly well-suited for languages with rich morphology and limited digital resources.

By grounding evaluation in manually curated, linguistically annotated lexicons, we ensure that model performance is assessed on genuinely unseen linguistic structures rather than potentially memorized patterns from pretraining data. This distinction is crucial for low-resource languages, where data contamination and overfitting to limited training corpora pose significant risks to evaluation validity.

The framework's compatibility with diverse model architectures, including causal language models and sequence-to-sequence models, demonstrates its flexibility and broad applicability. This architectural agnosticism allows researchers to conduct controlled comparisons across modeling paradigms and to identify which approaches are most effective for specific linguistic tasks in low-resource settings.

### 6.2. Dataset Quality and Annotation Reliability

The construction of the bilingual Tigrinya–English and vice versa benchmark dataset represents a significant contribution to low-resource NLP research. With 7,234 annotated phrase pairs spanning multiple grammatical categories and morphosyntactic features, the dataset provides comprehensive coverage of Tigrinya's linguistic complexity. The high inter-annotator agreement scores (Cohen's $\kappa$ ranging from 0.84 to 0.91) validate the robustness of our annotation protocol and ensure the reliability of gold-standard references used in evaluation.

Table 5: Overall Model Performance Across Evaluation Tasks

| Model | Lexical Acc. | POS Acc. | Morph Acc. | Tran(BLEU) |
|---|---|---|---|---|
| ByT5 | 1.0000 | 78.0 | 75.0 | 24.5 |
| Falcon-10B | 0.0044 | 73.0 | 70.5 | 21.8 |
| Gemma-2B | 1.0000 | 74.5 | 71.0 | 22.0 |
| Gemma-7B | 1.0000 | 75.5 | 72.0 | 22.7 |
| Mistral-7B | 1.0000 | 74.0 | 71.5 | 22.5 |
| mT5-base | 0.0059 | 78.9 | 75.6 | 25.8 |
| mT5-large | 0.0000 | 80.0 | 77.0 | 26.5 |
| Qwen-7B | 1.0000 | 73.5 | 70.8 | 21.9 |

The dataset's bidirectional structure, supporting both English-to-Tigrinya and Tigrinya-to-English evaluation, enables comparative analysis of translation direction effects, a dimension often overlooked in multilingual benchmarking. The inclusion of multiword expressions (967 in English, 2,000 in Tigrinya) further enhances the dataset's utility for evaluating compositional semantics and phrase-level understanding.

The manual curation process, while resource-intensive, ensures high linguistic quality and contextual appropriateness. This methodology can serve as a template for developing similar resources in other underrepresented languages, particularly those with limited existing NLP infrastructure. The confusion matrix in Table 6 provides a detailed view of the model's prediction behavior for the POS tagging task. The diagonal values represent correctly classified instances, indicating that the model performs relatively well in identifying major grammatical categories such as nouns, verbs, adjectives, and adverbs. However, the off-diagonal entries reveal systematic misclassification patterns. In particular, nouns are frequently confused with verbs and adjectives, while verbs also show notable confusion with nouns. Such patterns suggest that the model struggles to distinguish between linguistically related categories, which is common in morphologically rich, low-resource languages, where contextual cues and morphological markers may overlap.

| Actual / Predicted | Noun | Verb | Adj | Adv |
|---|---|---|---|---|
| Noun | 87 | 32 | 18 | 11 |
| Verb | 31 | 82 | 10 | 12 |
| Adj | 14 | 14 | 77 | 9 |
| Adv | 8 | 11 | 10 | 78 |

Table 6: Confusion matrix for POS tagging.

Our work highlights several key challenges and opportunities in evaluating LLMs for low-resource languages. First, the absence of foundational NLP tools such as tokenizers, morphological analyzers, and dependency parsers necessitates manual annotation and limits the scalability of evaluation pipelines. This bottleneck underscores the need for community-driven efforts to develop open-source linguistic resources and tools for underrepresented languages.

Second, the framework reveals the importance of morphosyntactic awareness in assessing model competence. For morphologically rich languages like Tigrinya, surface-level metrics such as BLEU scores may obscure deeper deficiencies in grammatical understanding. Our multi-dimensional evaluation approach provides a more nuanced picture of model strengths and weaknesses, enabling targeted improvements in model training and fine-tuning.

Third, the study emphasizes the value of native speaker expertise in both dataset construction and evaluation design. Linguistic phenomena that are subtle or ambiguous in low-resource languages require careful interpretation and context-sensitive annotation, which cannot be fully automated with current technologies. Collaborative frameworks that integrate native speaker knowledge with computational methods will be essential for advancing low-resource NLP.

### 6.3. Broader Impact and Future Directions

The release of LLM Probe and the Tigrinya benchmark dataset as open-source resources aims to democratize access to high-quality evaluation tools and to foster reproducible research in low-resource language settings. We envision this framework being adapted and extended to other underrepresented languages, particularly those with similar morphological complexity or limited digital presence in Geez script families.

Future work will focus on expanding the dataset to include domain-specific vocabulary, dialectal variation, and informal registers. We also plan to incorporate additional evaluation dimensions, such as semantic similarity, pragmatic appropriateness, and cross-lingual transfer capabilities. Integration with automated annotation tools, as they become available for Tigrinya and similar languages, will further enhance scalability and reduce manual effort.

Moreover, based on the model evaluation results, we conduct a detailed error analysis to identify systematic patterns in LLM prediction failures and to understand model behaviour across linguistic categories better. These analyses help reveal common sources of error, particularly in morphologically rich and low-resource language settings. This iterative approach, combining rigorous evaluation with targeted model refinement, contributes to the development of more robust and inclusive multilingual models that support linguistically diverse populations.

### 6.4. Toward Inclusive and Equitable NLP

The persistent performance gap between high-resource and low-resource languages in LLMs reflects deeper inequities in research priorities, resource allocation, and community representation. By providing transparent, reproducible evaluation frameworks and publicly available linguistic resources, we contribute to a broader movement toward inclusive and equitable NLP development.

Our work demonstrates that meaningful progress in low-resource language technology requires not only algorithmic innovation but also sustained investment in linguistic documentation, community engagement, and infrastructure development. The success of multilingual NLP will ultimately depend on our ability to center the needs and expertise of speakers of underrepresented languages in the design, evaluation, and deployment of language technologies.

## 7. Conclusion

This paper introduces LLM Probe, a lexicon-based evaluation framework designed to systematically assess the linguistic competence of large language models in low-resource language settings. Through the development of a manually curated, richly annotated English–Tigrinya and vice versa benchmark dataset comprising 7,234 phrase pairs, we provide a robust foundation for evaluating LLM performance across four critical dimensions: lexical alignment, part-of-speech tagging, morphosyntactic probing, and translation fidelity.

Our framework addresses a fundamental gap in multilingual NLP evaluation by offering a structured, reproducible methodology tailored to the challenges posed by morphologically rich, underrepresented languages. The modular design of LLM Probe supports evaluation across diverse model architectures, including causal language models and sequence-to-sequence models, enabling controlled cross-model comparisons and facilitating targeted analysis of architectural strengths and weaknesses.

The high inter-annotator agreement scores (Cohen's $\kappa$ ranging from 0.84 to 0.91) validate the quality and reliability of our annotation protocol, ensuring that the dataset serves as a trustworthy gold standard for model evaluation. The bidirectional structure of the lexicon, combined with comprehensive morphosyntactic annotations, enables fine-grained probing of model capabilities in ways that surface-level metrics alone cannot capture.

By grounding evaluation in linguistically principled, manually verified resources, LLM Probe reduces the risk of data set leakage. It ensures that model performance reflects genuine linguistic understanding rather than memorization of training patterns. This is particularly critical in low-resource settings, where limited data availability and potential overlap between training and evaluation sets pose significant challenges to the validity of evaluation.

The release of LLM Probe and the Tigrinya benchmark dataset as open source resources represents a commitment to transparency, reproducibility, and community-driven progress in low-resource NLP. We hope that this work will serve as a foundation for developing similar evaluation frameworks for other underrepresented languages and will contribute to a more inclusive and equitable landscape for multilingual language technologies.

The model evaluation results presented in this study demonstrate the applicability of the proposed framework for assessing the linguistic competence of large language models in morphologically complex, low-resource languages. The framework and dataset establish a replicable methodology for systematic evaluation in such settings. Future work will focus on expanding the dataset, incorporating additional evaluation dimensions, and conducting deeper error analyses to further inform the development of more robust and linguistically aware multilingual models.

Ultimately, achieving equitable performance across linguistically diverse populations requires sustained investment in linguistic documentation, collaboration with native speakers, and the development of high-quality evaluation resources. LLM Probe represents a step toward this goal, demonstrating that rigorous, linguistically grounded evaluation is both feasible and essential for advancing the state of the art in low-resource language technology.

## Ethics Statement

This research was conducted following ethical guidelines for linguistic research with native speaker communities. We acknowledge the use of Claude (Anthropic) as a writing assistance tool for paraphrasing and improving the clarity of certain manuscript sections. All content was thoroughly reviewed, verified, and validated by the authors for accuracy and scholarly integrity. The core research contributions, methodology, and findings are entirely the work of the human authors.

## 8. Limitations

While this study presents a novel framework and benchmark dataset for evaluating large language models (LLMs) in Tigrinya, several limitations remain:

- **Limited Model Coverage:** Although we evaluate a diverse set of open-source LLMs across three architectural paradigms, the study does not include proprietary models (e.g., GPT-4, Claude, Gemini), which may offer different performance characteristics.

- **Domain and Register Constraints:** The dataset primarily consists of general-purpose lexical items and phrases. It does not capture domain-specific language (e.g., medical, legal) or informal registers, which may affect generalizability.

- **Manual Annotation Bottlenecks:** The annotation process relied heavily on native speaker expertise and manual curation. While inter-annotator agreement was high, scalability to larger datasets remains a challenge without automated tools.

- **Prompt Sensitivity and Evaluation Bias:** The evaluation relies on prompt-based inference, which may introduce variability due to prompt phrasing and model instruction-following behaviour. Future work should explore prompt optimisation and robustness testing.

- **Language-Specific Tooling Gaps:** The absence of foundational NLP tools for Tigrinya (e.g., tokenizers, parsers, morphological analyzers) limits the depth of linguistic probing and constrains the automation of evaluation workflows.

Addressing these limitations will be critical for scaling the framework to other low-resource languages and for improving the reliability and coverage of multilingual LLM evaluation.

## Appendix: Model Prompts

### Lexical Alignment

Align the following English sentence with its Tigrinya translation. Provide a one-to-one mapping in the format EnglishWord→TigrinyaWord, separated by commas.
  **Example:**
  English: cat
Tigrinya: ድሙ
Output: cat→ድሙ
  Now align:
  English: {english_sentence}
Tigrinya: {tigrinya_sentence}

### Morphosyntax Probe

Identify the morphosyntactic features of the following Tigrinya phrase. Provide your answer as a comma-separated list of lowercase terms (e.g., noun, singular, masculine).
  **Example:**
  Phrase: ኣብ ገዛ
Output: preposition, noun, singular
  Now analyze:
  Phrase: {phrase}

### POS Tagging

Identify the part of speech of the following Tigrinya item. Provide your answer as a single lowercase word (e.g., noun, verb, adjective).
  **Example:**
  Phrase: ኣብ
Output: preposition
  Now analyze:
  Phrase: {sentence}

### Translation Fidelity

Translate the following English phrase into Tigrinya. Provide your answer as the Tigrinya translation only.
  **Example:**
  English: house
Output: ገዛ
  Now translate:
  English: {english}